\pdfoutput=1
\documentclass[runningheads]{llncs}
\usepackage[T1]{fontenc}
\usepackage{latexsym}
\usepackage{amssymb}
\usepackage{amsmath}
\usepackage{booktabs}
\usepackage{enumitem}
\usepackage{graphicx}
\usepackage{color}
\usepackage{multirow}
\usepackage{caption}
\usepackage{tabularx} 
\usepackage{graphicx}
\usepackage{subfigure}
\usepackage{caption}
\usepackage{latexsym}
\usepackage{graphicx}

\begin{document}
\title{The Master-Slave Encoder Model for Improving Patent Text Summarization: A New Approach to Combining Specifications and Claims}
\titlerunning{MSEA}
%
\author{Shu Zhou\inst{1,2}\thanks{These authors contributed equally.} \and
Xin Wang\inst{3} \and Zhengda Zhou\inst{1,2}$^{\star}$  \and Haohan Yi\inst{1,2}$^{\star}$ \and Xuhui Zheng\inst{1,2}  \and Hao Wang\inst{1,2}
\thanks{Corresponding author. This paper is supported by the National Natural Science Foundation of China under contract No. 72074108, Jiangsu Young Talents in Social Sciences, and Tang Schloar of Nanjing University.}
}

\authorrunning{S. Zhou et al.}
%
\institute{School of Information Management, Nanjing University,  China \and
Key Laboratory of Data Engineering and Knowledge Services in Jiangsu Provincial Universities, Nanjing University,  China\\
\email{shuzhou@smail.nju.edu.cn; ywhaowang@nju.edu.cn}\\
\and
Baidu, Inc., Beijing, China\\
}
\maketitle              

\begin{abstract}
In order to solve the problem of insufficient generation quality caused by traditional patent text abstract generation models only originating from patent specifications, the problem of new terminology OOV caused by rapid patent updates, and the problem of information redundancy caused by insufficient consideration of the high professionalism, accuracy, and uniqueness of patent texts, we proposes a patent text abstract generation model (MSEA) based on a master-slave encoder architecture; Firstly, the MSEA model designs a master-slave encoder, which combines the instructions in the patent text with the claims as input, and fully explores the characteristics and details between the two through the master-slave encoder; Then, the model enhances the consideration of new technical terms in the input sequence based on the pointer network, and further enhances the correlation with the input text by re weighing the "remembered" and "for-gotten" parts of the input sequence from the encoder; Finally, an enhanced repetition suppression mechanism for patent text was introduced to ensure accurate and non redundant abstracts generated. On a publicly available patent text dataset, compared to the state-of-the-art model, Improved Multi-Head Attention Mechanism (IMHAM), the MSEA model achieves an improvement of 0.006, 0.005, and 0.005 in Rouge-1, Rouge-2, and Rouge-L scores, respectively. MSEA leverages the characteristics of patent texts to effectively enhance the quality of patent text generation, demonstrating its advancement and effectiveness in the experiments.

\end{abstract}

\section{Introduction}
In recent years, as the number of patent applications in China has steadily increased, the quality of patent abstract generation has become increasingly scrutinized \cite{Zhang_2022_STNLTP}. Unfortunately, many patent applications are rejected, one major reason being defects in the claims, specifications, and abstracts \cite{Zhang_2021_RLCPAR}. These defects can partly be attributed to three key issues in abstract generation: first, the neglect of the importance of the claims; second, the Out of Vocabulary (OOV) issue with new patent terminology; and third, the failure to properly handle information redundancy and maintain professional value \cite{guoliang2023generating}.

Recently, encoder-decoder models based on neural networks have been used in various sequence-to-sequence tasks \cite{moratanch2016survey,syed2021survey}, such as machine translation, speech recognition, and text summary generation. Although these models have made significant progress in other sequence-to-sequence tasks like machine translation and speech recognition, they still face a range of challenges in patent summary generation \cite{Zhang_2022_STNLTP,guoliang2023generating}. We addresses the following three main issues:

Traditional patent text summary generation models \textbf{(1) only source input text from patent specifications}, resulting in summaries that may lack the core points of the patent. This is because patent texts include both the claims and the specifications, which differ in content (the specifications typically includes the title, technical field, background technology, content of the invention, specifications of the drawings, and specific embodiments, while the claims include independent and dependent claims) and importance.

The rapid updating of patents and the fine classification of technology result in a large number of new technological terms, \textbf{(2) making the OOV issue particularly prominent}. Because patents are updated quickly and classified in detail, new technological terms emerge easily, highlighting the OOV problem. However, traditional pointer networks often only consider the context vector, the state of the decoder, and the input to the decoder, and do not fully account for the importance of these new technological terms, leading to summaries that may neglect or mismanage these terms.

 The high professionalism and uniqueness of patent texts \textbf{(3) require summary generation models to be highly accurate and free of redundancy}. However, traditional coverage mechanisms have not fully addressed the challenge of repetitive generation, leading to summaries that may contain information redundancy and lose their professional value. Patent texts have a high degree of professionalism, precision, and uniqueness, where every specifications, term, and technical detail carries crucial information. Thus, compared to general text generation, repetitive generation in patent texts not only leads to information redundancy but also causes the generated text to lose its intended value and professionalism. Traditional coverage mechanisms have not fully adapted to and met the requirements for generating text content with such high specialization and uniqueness.

To address these issues, we proposes a new patent text summary generation model based on a master-slave encoder architecture (MSEA) on the existing sequence-to-sequence framework. MSEA considers the different importance of the specifications and claims in patent texts, dividing the encoder into two parts: a master encoder and a slave encoder, with a separate decoder. The slave encoder processes the input from the master encoder and other inputs separately, producing a new vector as an additional input to the decoder, which enables the decoder to obtain more semantic information. In the decoding phase, the paper conducts multi-step decoding operations and establishes a semantic feature vector for the text content at each step, allowing the decoder to continuously 'remember' the content generated in previous time steps to avoid repetition, thereby enhancing the accuracy and professional value of the summary.

The main contributions of this paper are summarized as follows:

(1) The master-slave encoder architecture (MSEA) is specifically designed to handle the specifications and claims in patent texts, integrating these two parts as inputs to fully explore their characteristics and details. This approach significantly improves the core point coverage of patent summaries, ensuring that the generated summaries accurately reflect the innovative aspects and technical scope of the patents.

(2) The MSEA model, through an improved pointer network and reweighted input sequence, particularly enhances the recognition and handling of new technological terms. This not only addresses the OOV issue but also enhances the model's relevance to the input text by 'remembering' and 'forgetting' different parts of the input sequence, improving the accuracy of term processing.

(3) The MSEA model introduces an enhanced suppression mechanism specifically tailored for the high professionalism and uniqueness of patent texts, ensuring that the generated summaries are both accurate and non-redundant. Through multi-step decoding and suppression of repetitive content, the model effectively prevents the generation of redundant information, ensuring the professional value and precision of the summaries.

(4) Experimental results show that the MSEA model surpasses advanced patent summary generation models (such as IMHAM) on common Rouge scoring metrics, demonstrating its advanced capabilities and effectiveness in the field of patent summary generation.

\section{Related work}
\subsection{Patent Text Abstract Generation}
Text summarization, particularly in the context of patent texts, remains a vital area of research in natural language processing. Deep learning advancements have notably improved summary quality, yet accurately summarizing main points continues to be a challenge \cite{moratanch2016survey}. Summarization techniques include extractive summarization, which compiles key sentences from the text \cite{nallapati2016abstractive}, and abstractive summarization, which creates summaries that may diverge from the source text to include multiple topics and varied sentence structures \cite{yang2020hierarchical}.

In patent summarization, existing models struggle with producing summaries of appropriate length and relevance \cite{moratanch2016survey}. Recent models like the Reinforcement Learning Chinese Patent Rewriting Abstract (RLCPRA) \cite{Zhang_2021_RLCPAR}, which focuses on patent specifications using reinforcement learning, and the Strategy Transformer Network Language Towards Patent (STNLTP) \cite{Zhang_2022_STNLTP}, which employs an ensemble approach for the same, address specific challenges such as out-of-vocabulary issues and repetitiveness. The Improved Multi-Head Attention Mechanism (IMHAM) \cite{guoliang2023generating} and Master-Slave Encoder Architecture (MSEA) models incorporate both specifications and claims of patents, aiming to enhance summary quality by acknowledging the unique structure of patent texts. MSEA differentiates itself by using a dual-decoder approach to capture more detailed aspects of patents compared to IMHAM's focus on either claims or specifications.

Despite advancements, combining specifications and claims in patent summarization remains an emerging field, indicating that the area is still evolving and underscoring the need for continued research.

\subsection{Out-of-vocabulary Problem}
The Out-Of-Vocabulary (OOV) issue is a significant challenge in natural language processing, where words not in a model’s vocabulary lead to errors in output. Recent strategies include subword tokenization \cite{kudo2018sentencepiece} and vocabulary construction with hierarchical supervision \cite{deng2019sentiment}, which allow models to generate words beyond their initial vocabulary, thus improving their handling of OOV words.

Additionally, incorporating external knowledge sources like word embeddings or dictionaries has shown to enhance text generation quality by providing extra context and meaning \cite{liu2020learning}. Moreover, the application of reinforcement learning  has furthered model performance by enabling iterative learning from feedback, optimizing the handling of OOV words in generated text.

\subsection{Repeated Generation Issues}
A common issue with neural network-based encoder-decoder models is their tendency to produce repetitive and incoherent phrases in longer summaries. To avoid this, a coverage mechanism can be used to eliminate repetitions by focusing on the same parts during encoding \cite{see2017get}. Additionally, the decoded information from the decoder can also be used to prevent repetition.

The problem of repetition occurs more frequently in long sequence generation tasks. However, researchers have seldom focused on using large datasets to summarize longer texts. Nallapati et al. \cite{nallapati2016abstractive} proposed an encoder-decoder model with hierarchical attention based on Recurrent Neural Networks (RNNs) for abstractive summarization tasks. Later, another hierarchical RNN model was developed, achieving significantly better results in abstraction \cite{nallapati2017summarunner}.


\section{Methodology}
In the task of generating patent text summaries, a patent typically contains information such as specification, claims, publication number, and title. This model specifically combines information from the specifications and claims. Specifically, it takes as input a sequence of source text from the patent's specifications \(X = (x_1, x_2, \ldots, x_j, \ldots x_m)\) and a sequence from the patent's claims \(X' = (x'_1, x'_2, \ldots, x'_j, \ldots x'_m)\), where \(j\) and \(m\) represent the index and number of words in the source text, respectively. The output is a summary sequence of the patent text \(Y = (y_1, y_2, \ldots, y_i, \ldots y_n)\), where \(i\) and \(n\) represent the index and number of words in the summary text, respectively.

\vspace{-10pt}
\begin{figure}[ht]
\centering
\includegraphics[width=1\linewidth]{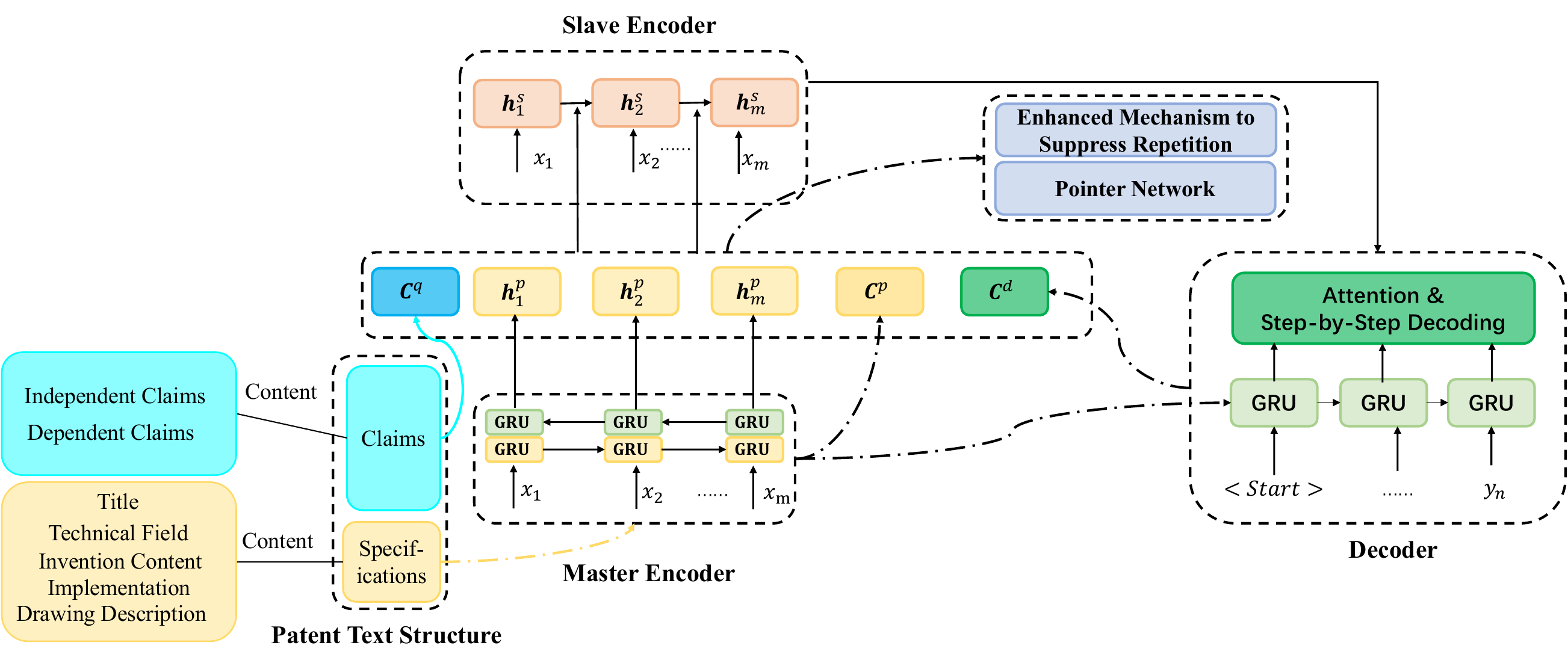}
\caption{The overall architecture of the model MSEA. MSEA has a master encoder, a slave encoder, and a decoder with an attention mechanism}
\label{fig:model}
\end{figure}
\vspace{-20pt}

This article will describe in detail the designed master-slave encoder model. The model structure is shown in Figure~\ref{fig:model}, consisting of a master encoder, a slave encoder, and a decoder with an attention mechanism:

At the master encoder end, the role of the master encoder is to calculate semantic vectors for each word in the input patent text's specifications.

At the slave encoder end, the slave encoder first calculates the importance weights of each word in the specifications of the input patent text, adds the text of each word from the claims of the patent text, and then recalculates the corresponding semantic vectors.

At the decoder end, a decoder with an attention mechanism is designed to decode in stages, producing a partially fixed-length output sequence at each stage.

\subsection{Master Encoder}
In this paper, GRU is used to adaptively capture dependencies between different time scales. As shown in Figure~\ref{fig:model}, the main encoder's relationships can be described by the following equations:
\begin{equation}
\begin{cases}
u_t = \sigma(W_u [x_t, h_{t-1}]) \\
r_t = \sigma(W_r [x_t, h_{t-1}]) \\
h_t' = \tanh(W_h [x_t, r_t \odot h_{t-1}]) \\
h_t = (1 - u_t) \odot h_{t-1} + u_t \odot h_t'
\end{cases}
\end{equation}
where \(W_u\), \(W_r\), and \(W_h\) are parameter matrices, and \(x_t\) and \(h_t\) represent the input vector and hidden state vector at time step \(t\), respectively.

The purpose of the main encoder is to construct feature representations of the input sentences from the patent specifications. In the recurrent unit part of the main encoder, Bi-GRU is used, which consists of a forward GRU and a backward GRU. Given an input sequence (such as \(x_1, x_2, \ldots, x_m\) in Figure 1), the forward GRU sequentially calculates the hidden state representations $\overrightarrow{h_1^p}, \overrightarrow{h_2^p}, \ldots, \overrightarrow{h_m^p}$ based on the current word embeddings. The backward GRU generates the hidden state representations \(\overleftarrow{h_1^p}, \overleftarrow{h_2^p}, \ldots, \overleftarrow{h_m^p}\) for each word in the reverse sequence. These two hidden states are defined as:
\begin{equation}
\begin{cases}
\overrightarrow{h_t^p} = \text{GRU}^p (x_t, \overrightarrow{h_{t-1}^p}) \\
\overleftarrow{h_t^p} = \text{GRU}^p (x_t, \overleftarrow{h_{t-1}^p})
\end{cases}
\end{equation}

The initial states of the Bi-GRU are set to zero vectors, i.e., \(\overrightarrow{h_1^p} = 0\) and \(\overleftarrow{h_1^p} = 0\). After the main encoder reads the input sentence, each word in the sentence can be represented by concatenating the forward and backward GRU states as \(h_t^p = [\overrightarrow{h_t^p}, \overleftarrow{h_t^p}]\). Subsequently, the representation of the text sequence in the patent specification is modeled as a nonlinear transformation of the average of the Bi-GRU hidden states. This is expressed as:
\begin{equation}
C^p = \tanh(W_p \cdot \frac{1}{N} \sum_{t=1}^{N} h_t^p + b_p)
\end{equation}
where \(W_p\) and \(b_p\) are hyperparameters, and \(N\) represents the length of the input sentence.

\subsection{Slave Encoder}

The slave encoder, depicted in Figure 1, utilizes a unidirectional GRU to process the input sequence from patent specifications every \(K\) decoding steps, creating hidden states and computing context via an attention mechanism. It calculates the importance weight \(\alpha_t\) based on the features of each word \(h_t^p\), the contents of the specifications \(C^p\), claims \(C^q\), and the decoder's output \(C^d\) as follows:
\begin{equation}
\begin{split}
    \alpha_t = \sigma(W_2 (\tanh(W_1 [h_t^p, C^p, C^q, C^d] + b_1)) + \\
    h_t^p{}^T W_s C^p + h_t^p{}^T W_s C^d - C^p{}^T W_r C^d + C^q{}^T W_k + b_2)\\
\end{split}
\end{equation}
where parameters \(W_1, W_2, W_s, W_r, W_k, b_1, b_2\) adjust how the model attends to and processes the input words \(x_t\) and \(x_t'\). 

The updating rule for the slave encoder's hidden state \(h_t^s\) based on \(\alpha_t\) is:
\begin{equation}
h_t^s = (1 - \alpha_t) \odot h_{t-1}^s + \alpha_t \odot \text{GRU}^s (x_t, h_{t-1}^s)
\end{equation}
this configuration allows the slave encoder to selectively integrate new input with prior information, aiding the decoder in generating accurate summaries by providing supplementary information through the final hidden state \(h_m^s\).

\subsection{Decoder}

In the task of generating patent text summaries, the slave encoder is used as a supplementary and dependency encoder to enhance the performance of the basic model of this paper. In the decoder part, a decoder with an attention mechanism is used, which calculates the context vectors based on the hidden states \(h_1^p, h_2^p, \ldots, h_m^p\) from the master encoder. The context vector \(c_i\) is calculated as the weighted sum of these hidden states, inspired by the Transformer, as shown in the following formula:
\begin{equation}
c_i = \sum_{j=1}^n a_{ij} h_j^p
\end{equation}
where each hidden state \(h_j^p\) weight \(a_{ij}\) can be computed as:
\begin{align}
a_{ij} &= \frac{\exp(e_{ij})}{\sum_{k=1}^n \exp(e_{ik})} \\
e_{ij} &= v_a^T \tanh(W_a h_{i-1}^d + U_a h_j^p) \\
h_i^d &= \text{GRU}^d(y_i, h_{i-1}^d)
\end{align}
Here, \(e_{ij}\) represents the matching degree of the input near position \(j\) with the output at position \(i\), and \(h_i^d\) is the hidden state generated by the decoder, which is based on its last hidden state \(h_{i-1}^d\) and the \(i\)th target in the output sequence \(y_i\).

The master-slave encoding model in this paper does not decode the entire output sequence at once, but decodes parts of a fixed length sequence in stages. Each stage decodes a part of the sequence of fixed length \(K\), and the entire decoded sequence is innovatively represented as:
\begin{equation}
C^d = \tanh(W_d \frac{1}{L} \sum_{i=1}^L h_i^p + b_d)
\end{equation}
where \(W_d\) and \(b_d\) are learning parameters, and \(L\) represents the length of the current coded sentence. \(C^d\) is the content currently produced by the decoder, which is used to adjust the attention weights of the slave encoder for each word in the input sequence.

After each fixed-length decoding, the slave encoder generates a new final state \(h_m^s\), and the decoder of this paper is innovatively rewritten as follows:
\begin{equation}
h_i^d = 
\begin{cases} 
\text{GRU}^d(y_i, [h_{i-1}^d, h_m^s]), & \text{if } L\%k == 0 \\
\text{GRU}^d(y_i, h_{i-1}^d), & \text{if } L\%k \neq 0
\end{cases}
\end{equation}
The initial state of the decoder is set to the final state of the master encoder's final state, i.e., \(h_0^d = h_m^p\). Every \(K\) decoding steps, the content after decoding and re-encoding is calculated. Then, the current context vector \(c_i\) obtained from the master encoder and the hidden state \(h_i^d\) of the decoder are concatenated and passed through a linear layer to produce a vocabulary distribution, as follows:
\begin{equation}
P_v = P(y_i \mid y_1, \ldots, y_{i-1}; x) = \text{softmax}(W_v [h_i^d, c_i] + b_v)
\end{equation}
where \(P(y_i \mid y_1, \ldots, y_{i-1}; x)\) is the conditional probability distribution of the target word \(y_i\) at time step \(i\), with \(W_v\) and \(b_v\) as learning parameters.

\subsection{Pointer Network}
This paper primarily utilizes a pointer network \cite{see2017get} to address the OOV problem. A soft switch \(P_p\) is used to choose between generating a word from a fixed vocabulary by sampling from \(P_v\) and copying a word from the input sequence by sampling from the attention distribution \(a_i\). \(P_p\) is the generation probability at time step \(i\), and the master formula is as follows:
\begin{equation}
P_p = \sigma(\omega_c^T c_i + \omega_h^T h_i^d + \omega_y^T y_i + \omega_d^T C^d + b_g)
\end{equation}
\begin{equation}
P_w = P_p P_v(w) + (1 - P_p) \sum_j^{w_i} a_{ij}
\end{equation}
where \(\omega_c^T\), \(\omega_h^T\), \(\omega_y^T\), \(\omega_d^T\), and \(b_g\) are learning parameters, \(c_i\) is the context vector, \(h_i^d\) is the hidden state of the decoder, \(y_i\) is the input to the decoder, \(C^d\) represents the content of the partially decoded sequence, and \(P_w\) is the probability distribution of the extended vocabulary. Details of the pointer network can be referred to in the original pointer network \cite{see2017get}.

\subsection{Enhanced Repetition Suppression Mechanism}

For sequence-to-sequence models, repetition is a common issue in sequence generation tasks, especially when generating multi-sentence texts. In this paper's model, an enhanced repetition suppression mechanism is adopted to address this problem. On one hand, the slave encoder generates an encoded feature vector every \(K\) steps, which enables the decoder to "remember" the content produced in earlier time steps to avoid repetition. On the other hand, this paper utilizes a coverage mechanism, where the coverage vector \(c^v\) is defined as the sum of attention distributions across all previous decoder time steps, innovatively expressed as:
\begin{equation}
c_i^v = \sum_{i'=0}^{i-1} a_{i'}
\end{equation}

Next, the coverage vector is also used as an additional input in the attention mechanism formula. Therefore, the formula for the attention mechanism is innovatively updated as:
\begin{equation}
e_{ij} = v_a^T \tanh(W_a h_{i-1}^d + U_a h_j^p + W_c h c_i^v)
\end{equation}

At the same time, we continue to define an additional coverage loss to penalize repetitive behavior. The formula for their loss function is written as:
\begin{equation}
L = \frac{1}{T} \sum_{i=0}^T \left(\Gamma_i + \lambda \sum_{j=0}^T \min(a_{ij}, c_{ij}^v)\right)
\end{equation}
where \(\lambda\) is a hyperparameter. \(i\) and \(j\) respectively represent the decoding time step and the position in the input sequence.

\section{Experiments}
\subsection{Data Source and Preprocessing}
The experimental data for this paper primarily comes from the Yizhuan Patent Retrieval and Analysis Database (https://www.patyee.com/), which includes patent data from most countries, available for download and research use. The data retrieval date range (publication date) is set from January 1, 2015, to January 1, 2022, covering China (including Hong Kong, Taiwan, and Macau). The patent types include invention applications, invention grants, utility models, design patents, and others, mainly covering five domains: water resources, artificial intelligence, fiber optics, finance, and agriculture. The patent status is set to valid, and the patent language is Chinese, selecting a total of 50,769 water resource patents, 32,939 artificial intelligence patents, 126,987 fiber optics patents, 18,758 finance patents, and 36,483 agriculture patents.

For data processing, this paper uses regular expressions to remove special characters, punctuation, spaces, and other special formats. References to images and other information contained in the patent claims and specifications also need to be removed, along with web tags and image references. The types of regular expression processing and corresponding expressions are shown in Table~\ref{tbl:process}:

\vspace{-25pt}
\begin{table}[h]
\caption{Regular expression processing types and their expressions}
\centering
\begin{tabular}{ll}
\toprule
\textbf{Regular Expression} & \textbf{Function} \\ 
\midrule
\texttt{<script[\^{}>]*?>[\textbackslash{}s\textbackslash{}S]*?<\textbackslash{}/script>} & Handles web tags and similar formats \\ 
\texttt{<style[\^{}>]*?>[\textbackslash{}s\textbackslash{}S]*?<\textbackslash{}/style>} &  \\ 
\texttt{<(?!div|/div|p|/p|br)[\^{}>]*>} &  \\ 
\texttt{<tr>(.*?)</tr>} &  \\ 
\texttt{<th>(.*?)</th>} &  \\ 
\texttt{<td>(.*?)</td>} &  \\ 
\texttt{(?<=<title>).*?(?=<\textbackslash{}/title>)} & Processes titles \\ 
\texttt{<a.*?href=.*?<\textbackslash{}/a>} & Handles image references and hyperlinks \\ 
\texttt{\textbackslash{}s*|\textbackslash{}t|\textbackslash{}r|\textbackslash{}n} & Handles excess spaces and lines \\ 
\bottomrule
\end{tabular}
\label{tbl:process}
\end{table}
\vspace{-15pt}

Data is split into training, testing, and validation sets at ratios of 8:1:1. Each downloaded data entry includes a title, publication number, abstract, specification text, and claims. An example of the patent data presentation form for the artificial intelligence domain is shown in Table~\ref{tbl:exampleofpatent}:

\begin{table}[h]
\caption{An Example of a Patent Data Presentation Form}
\label{tbl:exampleofpatent}
\centering
\begin{tabular}{p{0.15\linewidth} p{0.8\linewidth}}
\hline
\textbf{Attribute} & \textbf{Content} \\
\hline
Title & Task Scheduling Method and Device Based on Multiple GPUs \\ 
\hline
Publication Number & CN113391905A \\ 
\hline
Abstract & The invention discloses a task scheduling method and device based on multiple GPUs, including: allocating a minimum and maximum number of GPUs for different task types; loading tasks from the database into the task queue, distributing GPUs ... \\ 
\hline
Specification Text & The invention aims to provide a task scheduling method and device based on multiple GPUs. Technical solution: The invention provides a task scheduling method that includes: determining the priority of task types, allocating a minimum and maximum number of GPUs for different task types... \\ 
\hline
Claims & The second scheduling unit is for, if the number of GPUs in use has reached the minimum GPU requirement for each task type, or all tasks of a task type have been satisfied, and there are still tasks in the task queue and available GPUs... \\ 
\hline
\end{tabular}
\end{table}
\vspace{-45pt}

\subsection{Model parameter settings and Metrics}
For model training, the batch training size is set to 32, the probability of DROPOUT is set to 0.5, the initial learning rate is set to 0.001, the master encoder hidden layer dimension is set to 256, and the slave encoder hidden layer dimension is set to 256, the decoder hidden layer dimension setting  is set to 256, the maximum text input length  is set to 500, the maximum text output length  is set to 100 and the optimizer uses Adam. the maximum vocabulary size  is set to 100000. the OOV text is replaced with <UNK>. In this paper, we do not pre-empt word embeddings, but learn them from scratch in training. The dimension of word embeddings is set to 256. network parameters are randomly initialized over a uniform distribution [-0.05,0.05]. Every $K$ steps from the encoder was set to 100.

In the field of text generation, the system-generated abstracts are compared with the manual abstracts of the patents themselves, and the specific quality of their text generation is evaluated by calculating their overlap. In this paper, ROUGE is used as a measure of the model's effectiveness.

\subsection{Baselines}
In this paper, we compare with other current models in the same field, SuRuNNer \cite{nallapati2017summarunner} based on a recurrent neural network sequence model and combined with an attention mechanism \cite{see2017get} for extractive summarization. TextRank \cite{mihalcea2004textrank} uses a graph-based text processing model with 2 innovative unsupervised keyword extraction methods. MedWriter \cite{pan2020medwriter} uses a knowledge-aware model of text generation with a capability to learn graph-level representations illustrated. RLCPRA \cite{Zhang_2021_RLCPAR} uses reinforcement learning for patent text generation. STNLTP \cite{Zhang_2022_STNLTP} uses an integrated strategy for text generation. IMHAM \cite{guoliang2023generating} applies improved multi-head attention to decoders and encoders using most important document semantic similarity selection and pointer network optimization, which belongs to the current state-of-the-art of more advanced patent generating summary generation models.

\subsection{Comparison}
As shown in Table~\ref{tbl:compare}, the specific evaluation metrics for each model on ROUGE are displayed. It can be observed that MSEA performs better than TextRank, SuRuNNer, MedWriter, and IMHAM across all metrics. On Rouge-1, Rouge-2, and Rouge-L, MSEA scores higher by 0.006, 0.005, and 0.005 respectively compared to the currently advanced IMHAM, and by as much as 0.109, 0.117, and 0.08 compared to TextRank. However, the performance of Transformer was not as good as MedWriter and IMHAM, possibly because IMHAM and MedWriter incorporate more features related to patent structure and use Bert. The reason MSEA performs better than RLCPRA might be that RLCPRA uses reinforcement learning to generate text specifically for the patent specifications to address OOV and repetitive generation issues. MSEA performs better than STNLTP possibly because STNLTP employs an integration strategy solely for generating text from the patent specifications. The experiments demonstrate that MSEA consistently shows the best results across all metrics, indicating that the model in this paper has achieved good performance.

\vspace{-25pt}

\begin{table}[h]
\centering
\caption{ROUGE performance of the models on the patent text dataset}
\begin{tabular}{lccc}
\toprule
\textbf{Model} & \textbf{Rouge-1} & \textbf{Rouge-2} & \textbf{Rouge-L} \\ 
\midrule
TextRank \cite{mihalcea2004textrank} & 0.432 & 0.235 & 0.367 \\ 
SuRuNNer (2017) \cite{nallapati2017summarunner} & 0.482 & 0.293 & 0.393 \\ 
Transformer (2017) & 0.491 & 0.316 & 0.402 \\ 
MedWriter (2020) \cite{pan2020medwriter} & 0.518 & 0.339 & 0.419 \\ 
RLCPRA (2021) \cite{Zhang_2021_RLCPAR} & 0.521 & 0.342 & 0.424 \\ 
STNLTP (2022) \cite{Zhang_2022_STNLTP} & 0.528 & 0.344 & 0.431 \\ 
IMHAM (2023 )\cite{guoliang2023generating} & 0.535 & 0.347 & 0.442 \\ 
MSEA & 0.541 & 0.352 & 0.447 \\ 
\bottomrule
\end{tabular}
\label{tbl:compare}
\end{table}
\vspace{-30pt}

\subsection{Analysis of Differences between Using Specifications and Claims Text in Patents}
As shown in Table~\ref{tbl:usesorc}, the differences in the performance of the MSEA model when using the specifications text and the claims text under various conditions are presented. The MSEA model achieves the best results when both the specifications and claims texts are used together. When only the specifications text or the claims text is used, the results are not as good as when both are used simultaneously, which also validates the superiority of the MSEA model.

\begin{table}[h]
\centering
\caption{ROUGE performance of the model on whether or not to use the text of the patent specifications and claims}
\begin{tabular}{lccc}
\toprule
\textbf{Use of specifications or Claims Text} & \textbf{Rouge-1} & \textbf{Rouge-2} & \textbf{Rouge-L} \\ 
\midrule
specifications Only & 0.537 & 0.348 & 0.443 \\ 
Claims Only & 0.534 & 0.345 & 0.444 \\ 
Both Used & 0.541 & 0.352 & 0.447 \\ 
\bottomrule
\end{tabular}
\label{tbl:usesorc}
\end{table}

\subsection{Sensitivity Analysis Under Different Decoding Lengths}

In the MSEA model designed in this paper, to assess the impact of different decoding lengths on performance, various decoding lengths \(K = \{20, 30, 50, 100, 150, 200\}\) were set. A decoding length of 200 means the entire output sequence can be decoded in one go. As shown in Table~\ref{difflength}, the ROUGE scores of the model at different decoding lengths are displayed.

\vspace{-25pt}
\begin{table}[h]
\centering
\caption{ROUGE performance of different decoding lengths on the patent text dataset}
\begin{tabular}{cccc}
\toprule
\textbf{Decoding Length} & \textbf{Rouge-1} & \textbf{Rouge-2} & \textbf{Rouge-L} \\ 
\midrule
20 & 0.533 & 0.343 & 0.437 \\ 
30 & 0.536 & 0.347 & 0.444 \\ 
50 & 0.538 & 0.346 & 0.443 \\ 
100 & 0.540 & 0.349 & 0.446 \\ 
150 & 0.537 & 0.351 & 0.441 \\ 
200 & 0.531 & 0.343 & 0.435 \\ 
\bottomrule
\end{tabular}
\label{difflength}
\end{table}\
\vspace{-30pt}

As can be seen from Table~\ref{difflength}, performance significantly decreases when the decoding length is too short. Setting the decoding length between 100 and 150 results in better outcomes.

\subsection{Sensitivity Analysis of Hidden Layers}

The settings of the hidden layer dimensions for the master encoder, slave encoder , and decoder significantly impact experimental results, as these layers contain crucial feature information. As shown in Table 6, the effects of the hidden layer sizes for the master encoder, slave encoder, and decoder on experimental outcomes are presented.

\vspace{-25pt}
\begin{table}[h]
\centering
\caption{Effect of hidden layer size of master encoder, slave encoder, and decoder on experimental metrics Rouge-1, Rouge-2, and Rouge-L}
\begin{tabular}{cccccccccc}
\toprule
\textbf{Rouge Metric} & \multicolumn{3}{c}{\textbf{Master Encoder}} & \multicolumn{3}{c}{\textbf{Slave Encoder}} & \multicolumn{3}{c}{\textbf{Decoder}} \\ 
\midrule
& 128 & 256 & 512 & 128 & 256 & 512 & 128 & 256 & 512 \\ 
\textbf{Rouge-1} & 0.535 & 0.537 & 0.532 & 0.537 & 0.541 & 0.534 & 0.539 & 0.541 & 0.535 \\ 
\textbf{Rouge-2} & 0.348 & 0.349 & 0.349 & 0.347 & 0.352 & 0.348 & 0.351 & 0.351 & 0.348 \\ 
\textbf{Rouge-L} & 0.442 & 0.443 & 0.439 & 0.441 & 0.446 & 0.441 & 0.445 & 0.445 & 0.440 \\ 
\bottomrule
\end{tabular}
\label{tbl:effectlayer}
\end{table}
\vspace{-20pt}

From Table~\ref{tbl:effectlayer}, it is observed that the model achieves optimal results on the Rouge-1, Rouge-2, and Rouge-L metrics when the hidden layer sizes are set to 256 for the master encoder, slave encoder, and decoder. This indicates that both excessively large or small hidden layers do not favor good model performance, while moderately sized hidden layers help further enhance performance.

\subsection{Ablation Study}

To further explore the impact of different modules on the experimental results, an ablation study was conducted. Specifically, one or two of the three modules were removed from the MSEA model each time. As shown in Table~\ref{tbl:ablation}, particularly, in the absence of both the master-slave encoding mechanism and all modules, the model reverts to a classic sequence-to-sequence model.

\vspace{-25pt}
\begin{table}[h]
\centering
\caption{Experimental analysis of ablation of the MSEA model}
\begin{tabularx}{\textwidth}{Xccc} 
\hline
\textbf{Variable} & \textbf{Rouge-1} & \textbf{Rouge-2} & \textbf{Rouge-L} \\ \hline
Model Itself & 0.541 & 0.352 & 0.447 \\ 
No Enhanced Repetition Suppression & 0.524 & 0.337 & 0.423 \\ 
No Master-Slave Encoding & 0.534 & 0.346 & 0.434 \\ 
No Pointer Network & 0.538 & 0.349 & 0.438 \\ 
No Pointer Network + No Enhanced Repetition Suppression & 0.478 & 0.276 & 0.382 \\ 
No Master-Slave Encoding + No Enhanced Repetition Suppression & 0.512 & 0.309 & 0.408 \\ 
No Enhanced Repetition Suppression + No Master-Slave Encoding + No Pointer Network & 0.466 & 0.268 & 0.377 \\ \hline
\end{tabularx}
\label{tbl:ablation}
\end{table}
\vspace{-20pt}

From Table~\ref{tbl:ablation}, it is evident that the absence of the repetition suppression mechanism leads to the largest drop in ROUGE scores, indicating that repetitive phenomena significantly affect the performance of summary generation, and the "RA" of this paper can effectively suppress such repetitions. The absence of the master-slave encoding mechanism results in decreases of 0.007, 0.006, and 0.013 in Rouge-1, Rouge-2, and Rouge-L, respectively. This suggests that the slave encoder performs more precise encoding, aiding the model in considering more detailed information. A performance decline is also noted without the pointer mechanism, likely due to the increased appearance of OOV words in the generated summaries. Thus, these three components are crucial for the performance of the MSEA model.

\subsection{Enhanced Repetition Suppression Mechanism Results Analysis}
The master-slave encoding model in this paper utilizes an enhanced repetition suppression mechanism that integrates the existing coverage mechanism with outputs already generated by the decoder. To verify the capability of the summary generated without the coverage mechanism to eliminate repetitive phenomena, the decoding length was set to a smaller value to enable the decoder to "remember" decoding information from earlier time steps better.

\begin{table}[h]
\centering
\caption{Case analysis for different decoding lengths and their effect on output quality}
\label{tbl:case}
\scalebox{0.9}{
\begin{tabular}{p{0.1\linewidth} p{0.9\linewidth}}
\hline
\textbf{Model} & \textbf{Summary Content} \\ \hline
Original Content & Hydraulics engineering projects undertake tasks of water retention and drainage, thus requiring special properties such as stability, pressure bearing, impermeability, abrasion resistance, frost resistance, and crack resistance in hydraulic structures. According to the technical specifications of hydraulic engineering, specific construction methods and measures must be taken to ensure the quality of the work. However, in practical use, embankments are easily loosened after prolonged exposure to tidal impacts, posing significant risks, and the water surface at the embankment carries a lot of floating large debris, which is very inconvenient for workers to salvage. \\ \hline
Manual Summary & This utility model discloses a water engineering anti-surge embankment protection device, related to the technical field of water engineering. The device includes two support rods, two second connection plates, and two first connection plates, all fixed at both ends of the top of the protective board. The device features a protection cleaning mechanism, sliding of the reinforcement plate within the groove to facilitate adjustment of its position for adapting to different water levels, rotation of the reel to adjust the position of the collection plate, facilitating workers in collecting surface garbage, with a simple overall design and compact structure, which protects the embankment while facilitating cleaning of surface garbage, possessing good practicality. \\ \hline
MSEA, \(K=200\) & This utility model relates to the technical field of water engineering, specifically an embankment protection device for water engineering, including a protective board with two support rods, two second connection plates fixed at both ends of the top of the protective board. The protective board's surface has two grooves, and the protective cleaning mechanism is set on the surface. The protective cleaning mechanism includes a reinforcement plate, and the support plate is set on the surface of the protective board, able to adjust... \\ \hline
MSEA, \(K=150\) & This utility model pertains to the technical field of water engineering, especially an embankment protection device for water engineering, including a protective board with two support rods and two second connection plates fixed at both ends of the top of the protective board. The device features a cleaning part that facilitates adjustment of the reinforcement plate, suitable for different water levels, easy position adjustment, and convenient garbage cleaning. This device is ingeniously designed for embankment protection and is useful. \\ \hline
\end{tabular}
}
\end{table}

Table~\ref{tbl:case} displays examples generated by the MSEA model of this paper without using the coverage mechanism. Results show that when the decoding length is set to a smaller value, the master-slave encoding model without the coverage mechanism still manages to suppress repetition.

\section{Conclusions}

This study enhanced patent text summarization by developing a master-slave encoder architecture (MSEA) model that integrates patent specifications and claims, significantly improving summary quality. The MSEA model addresses traditional limitations by incorporating a pointer network for handling new technological terms and an enhanced repetition suppression mechanism to reduce content redundancy, thereby overcoming the inadequacies of previous summarization methods and coverage mechanisms.
%
%
%
\bibliographystyle{splncs04}
\bibliography{mybibfile}
%




\end{document}